\documentclass[letterpaper, 10 pt, conference]{ieeeconf}  

\IEEEoverridecommandlockouts                              

\usepackage{graphics} 
\usepackage{epsfig} 
\usepackage{mathptmx} 
\usepackage{times} 
\usepackage{amsmath} 
\usepackage{amssymb}  
\usepackage{booktabs}
\usepackage{multirow} 
\usepackage{url}
\usepackage{hyperref}
\usepackage{cite}
\usepackage{xcolor}
\usepackage{graphicx}
\usepackage{makecell}
\DeclareGraphicsExtensions{.jpg, .png}
\title{\LARGE \bf
ADP: Adversarial Dynamics Priors for Physically Grounded\\Humanoid Locomotion
}

\author{Anonymous Author(s)}%

\author{Seokju Lee$^{1*}$, Jeongtae Lee$^{1*}$, Jeonghyeok Lim$^{1}$, Jeonguk Kang$^{2}$, Byungwook Lee$^{1}$, Seungho Han$^{3}$, \\Keun Ha Choi$^{1}$, Dongil Park$^{4,\dagger}$, and Kyung-Soo Kim$^{1,\dagger}$
\thanks{*Equal contribution.}
\thanks{This work was supported by R\&D Project of Korea Institute of Machinery and Materials (No. NK261A). \textit{(Corresponding author: Dongil Park and Kyung-Soo Kim)}}
\thanks{$^{1}$Seokju Lee, Jeongtae Lee, Jeonghyeok Lim, Byungwook Lee, Keun Ha Choi, Kyung-Soo Kim are with the Mechatronics, Systems and Control Lab (MSC Lab), Department of Mechanical Engineering, Korea Advanced Institute of Science and Technology (KAIST), Yuseong-gu, Daejeon 34141, Republic of Korea (e-mail: \{dltjrwn0322, us03316, henricus0973, bwlee605, choiha99, kyungsookim\}@kaist.ac.kr)}
\thanks{$^{2}$Jeonguk Kang is with the Samsung Electronics, Future Robotics AI Group, Seoul, Republic of Korea. (e-mail: kju8765@gmail.com)}
\thanks{$^{3}$Seungho Han is with the School of Electrical Engineering, Hanyang University, Ansan 15588, Republic of Korea. (email: seunghohan@hanyang.ac.kr)}
\thanks{$^{4}$Dongil Park is with the Advanced Robotics Research Center, Korea Institute of Machinery \& Materials (KIMM), Daejeon 34103, Republic of Korea. (e-mail: parkstar@kimm.re.kr)}
\thanks{Paper website: \url{https://seokju-lee.github.io/adp}}}%

\begin{document}

\maketitle
\thispagestyle{empty}
\pagestyle{empty}

\begin{abstract}
In this paper, we propose Adversarial Dynamics Priors (ADP) for perturbation-resilient humanoid locomotion control. Existing motion prior-based methods induce natural motion styles by imitating kinematic motion features, but they do not directly regularize dynamics features, such as CoM motion, centroidal momentum, contact forces, and contact states. To address this limitation, we replace kinematic motion-style feature with selected dynamics features extracted from locomotion trajectories as the target of adversarial regularization. To this end, we use trajectory optimization to construct a reference dataset and train a discriminator to evaluate whether policy-induced temporal windows are consistent with the resulting reference distribution. Without explicit motion tracking, ADP encourages policy rollouts to remain close to the reference support, even after perturbations. Experimental results show that, compared with AMP, the strongest baseline in our evaluation, ADP improves the $80\%$-success impulse threshold ($J_{80}$) by $16.7\%$, while reducing direction-averaged recovery time and velocity tracking error by $47.9\%$ and $35.4\%$, respectively.
\end{abstract}

\section{INTRODUCTION}
Humanoid robots have recently attracted increasing attention as robotic platforms capable of operating in human-centric environments and performing tasks on behalf of humans~\cite{ze2025generalizable, liu2025opt2skill, fu2026demohlm}. For controlling such high-dimensional systems, learning-based control has become a dominant paradigm, building on its remarkable success in quadrupedal locomotion~\cite{lee2020learning, miki2022learning, choi2023learning}. Nevertheless, humanoid locomotion remains substantially more challenging than quadrupedal locomotion due to the increased number of degrees of freedom, the underactuated floating-base dynamics, and the need to maintain balance with intermittent contacts. Consequently, recent approaches have moved beyond purely hand-designed reward functions and have instead leveraged motion datasets to guide policy learning, either through explicit reference-tracking rewards or through motion priors~\cite{HeT-RSS-25, allshire2025visual, liao2025beyondmimic, sleiman2026zest}.

A representative approach for leveraging motion datasets is to design an imitation reward that directly tracks a reference motion. For example, DeepMimic~\cite{peng2018deepmimic} trains a policy to follow a given reference motion using tracking terms such as pose, velocity, end-effector, and root-state matching. Although this tracking-based formulation is effective for generating natural motions, its reliance on phase variables or target poses can limit policy flexibility when external perturbations drive the robot away from the reference trajectory. To mitigate this limitation, AMP~\cite{peng2021amp} formulates motion imitation as a distribution matching problem rather than explicit tracking, and uses a discriminator-based style reward to encourage the policy to generate a kinematic motion-style distribution similar to the motion dataset. However, the AMP prior is still based on kinematic motion features such as joint pose, velocity, and end-effector state, and therefore does not directly regularize dynamics-level recovery behaviors required after perturbations, such as momentum regulation, contact-force modulation, and contact timing.

\begin{figure}
    \centering
    \includegraphics[width=\linewidth]{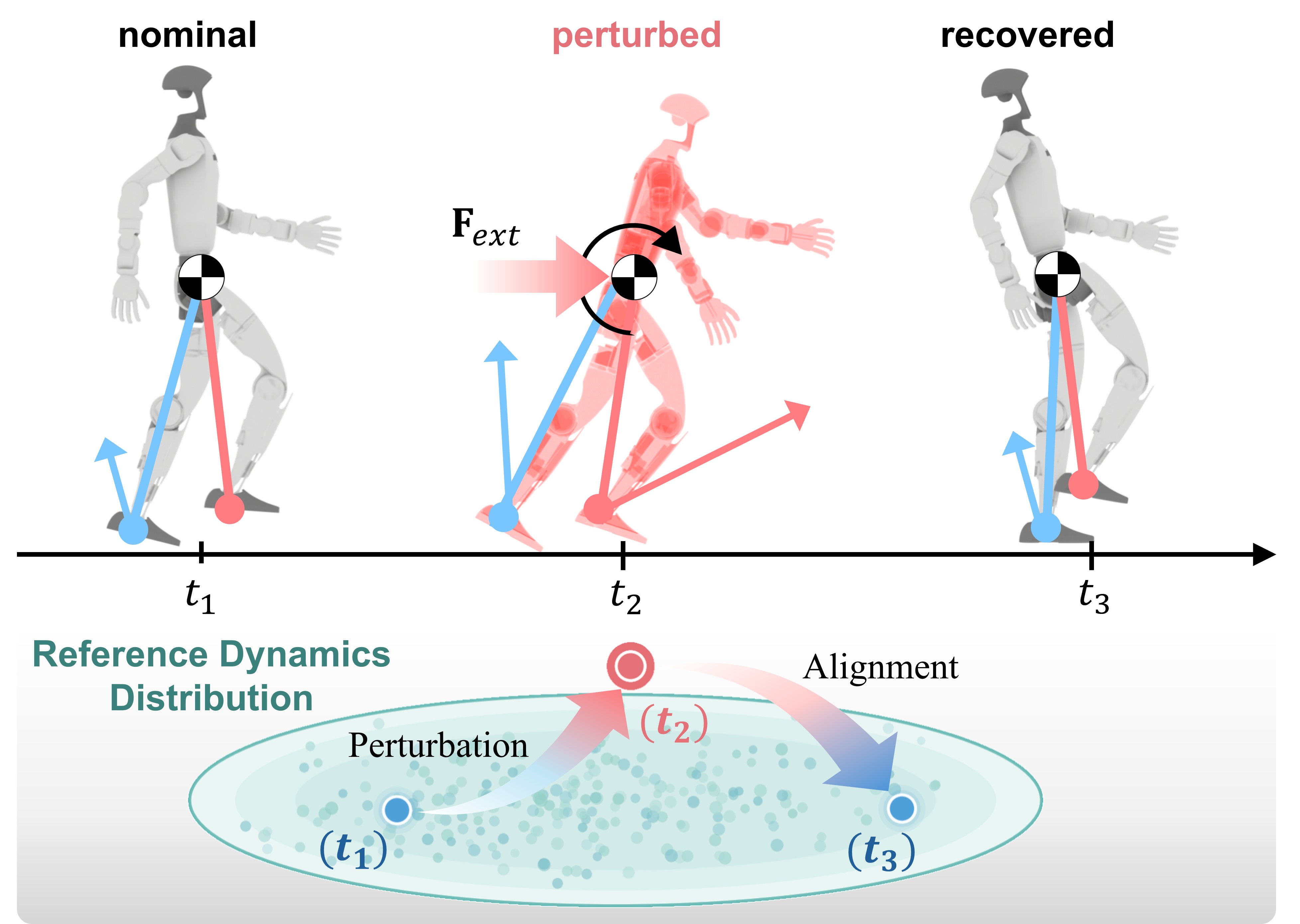}
    \caption{Dynamics-feature alignment for perturbation recovery. An external force at $t_2$ perturbs the humanoid and drives the generated dynamics features away from the reference distribution. ADP encourages recovery by regularizing the perturbed windows toward this distribution, resulting in stable locomotion at $t_3$.}
    \label{fig:figure1}
\end{figure}


In this paper, we propose Adversarial Dynamics Priors (ADP), which leverages dynamics features obtained from Trajectory Optimization (TO) to address the limitations of motion-style priors. Instead of directly imitating the kinematic style of reference motions, ADP provides an adversarial reward that keeps policy-induced dynamics features close to a reference distribution. We construct $\mathcal{D}_{\mathrm{dyn}}$ by solving a Single-Rigid-Body-Dynamics (SRBD)-based TO problem and extracting dynamics features from the resulting locomotion trajectories. The resulting samples satisfy SRBD-level dynamics and contact constraints, forming the reference set used for adversarial training. During policy training, the same feature type is computed from rollouts, and a discriminator is trained to distinguish reference windows from policy-generated windows. Conversely, the policy is optimized to maximize both the task reward and the discriminator-based dynamics prior reward, thereby learning recovery behaviors that remain close to the reference distribution even after external perturbations, as shown in Fig.~\ref{fig:figure1}. The main contributions of this paper are as follows:

\begin{itemize}
    \item \textbf{Adversarial Dynamics Priors:} We propose ADP, which replaces kinematic motion-style features with selected dynamics features for adversarial regularization, including CoM motion, centroidal momentum, contact forces, and contact states, without reference pose, phase, or end-effector tracking.
    \item \textbf{Perturbation-Sensitive Dynamics Representation:} We formulate a temporal dynamics-feature representation that exposes post-push transients in CoM motion, centroidal momentum, contact forces, and contact timing more directly than joint-level kinematic features.
    \item \textbf{Perturbation Recovery Evaluation:} We show through simulation comparisons with Vanilla RL, AMP, and direct dynamics-feature rewards that ADP improves humanoid perturbation recovery, and provide qualitative hardware demonstrations on a real humanoid robot under external perturbations.
\end{itemize}

The remainder of this paper is organized as follows. Section~\ref{section:related_work} reviews prior studies on learning policy networks using motion datasets and dynamics datasets. Section~\ref{section:method} describes the proposed method in detail. Section~\ref{section:experiments} presents the experimental setup and results. Finally, Section~\ref{section:conclusion} concludes the paper and discusses future directions.

\section{RELATED WORK}
\label{section:related_work}
\subsection{Motion Imitation for Robot Control}


Recent robot control has widely adopted motion imitation methods to learn natural behaviors for complex articulated systems. In particular, for legged robots, designing hand-designed rewards alone is often insufficient to induce natural locomotion, and many studies have leveraged motion datasets from animals or humans to assist policy learning. Escontrela $\textit{et al.}$~\cite{escontrela2022adversarial} replaced complex locomotion reward design with an AMP-based method that learns an adversarial style reward from a motion dataset, demonstrating natural and energy-efficient locomotion on a quadruped robot. Wu $\textit{et al.}$~\cite{wu2023learning} further used a motion dataset generated by SRBD-based TO to learn an AMP style reward, and combined it with task and regularization rewards to achieve robust and agile quadruped locomotion over diverse terrains.

These motion imitation approaches have also become an important direction for humanoid robot control. Since humanoids have high degrees of freedom and unstable floating-base dynamics, it is difficult to learn natural and stable whole-body behaviors using simple task rewards alone. Accordingly, many studies explicitly track kinematic features of reference motions, such as joint positions, joint velocities, root states, and end-effector positions, or use adversarial imitation learning to match kinematic motion-style distributions~\cite{tang2024humanmimic}. While these methods are effective in generating human-like natural behaviors, their priors are still largely based on kinematic motion features. However, perturbation recovery requires not only kinematic motion-style similarity but also appropriate momentum regulation and contact interaction for restoring balance. This motivates replacing kinematic motion-style matching with dynamics-level adversarial regularization.

To address this limitation, ADP replaces kinematic motion imitation with adversarial dynamics-feature regularization, encouraging policy-generated dynamics features to remain close to a reference dynamics-feature distribution even after external perturbations.
\subsection{Dynamics-Guided Locomotion Learning}
\begin{figure*}
    \centering
    \includegraphics[width=\linewidth]{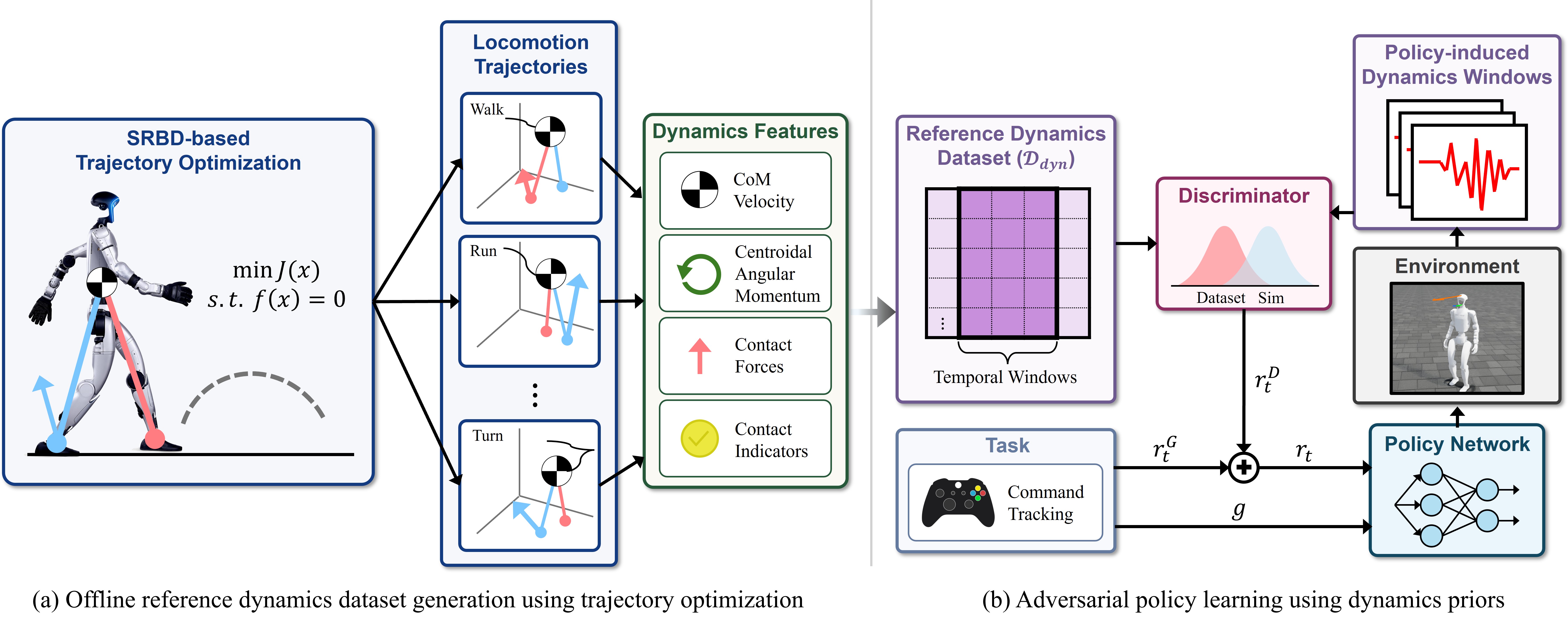}
    \caption{Overview of the proposed Adversarial Dynamics Priors (ADP) framework. (a) SRBD-based trajectory optimization generates reference locomotion trajectories, from which dynamics features are extracted to construct $\mathcal{D}_{\mathrm{dyn}}$. (b) During policy training, policy-generated windows are evaluated by a discriminator trained with $\mathcal{D}_{\mathrm{dyn}}$, producing the dynamics prior reward $r_t^D$. The policy is trained with the task reward $r_t^G$, dynamics prior reward $r_t^D$, and regularization reward $r_t^R$. }
    \label{fig:overview}
\end{figure*}

Dynamics-based control has long provided a physical basis for stable legged locomotion. Centroidal dynamics represents complex multibody motion through CoM motion and centroidal momentum, and has been widely used for humanoid motion planning and balance control~\cite{orin2013centroidal}. Building on this representation, WBC, MPC, and TO methods incorporate full-body kinematics, contact constraints, and friction constraints to generate stable and dynamic motions for humanoid and quadruped robots~\cite{dai2014whole,dafarra2022dynamic,winkler2017fast,farshidian2017robust,neunert2017trajectory,kim2018computationally,kim2019highly}. Related approaches have further used DCM or ZMP reference adaptation for push and stumble recovery, compliant walking-pattern generation, and whole-body motion generation under inertia constraints~\cite{mesesan2021online,kim2019online,ficht2020fast}. However, these methods often depend on explicit dynamics models, contact-state estimation, prescribed contact schedules, online optimization, or separate tracking controllers, which can limit rapid feedback adaptation under model mismatch or unexpected contact changes.

To mitigate these limitations, recent studies have attempted to incorporate the physical structure of model-based dynamics into learning-based locomotion. For example, TO-generated demonstrations have been used to alleviate the initial exploration problem in RL~\cite{bogdanovic2022model}, and LIP-dynamics-based footstep guidance has been incorporated into RL policy learning~\cite{lee2024integrating}. Other approaches have used contact-consistent whole-body trajectories generated from full-order dynamics-based TO as imitation targets~\cite{liu2025opt2skill}, or embedded a differentiable simulator into a generative model to produce physically consistent state-action trajectories~\cite{lee2025dynaflow}. Although DynaFlow ensures the dynamic feasibility of generated trajectories, its dynamics model is embedded within the generative process to synthesize feasible motions, rather than used as a prior for evaluating the closed-loop dynamics produced by an RL policy. More recently, APT-RL demonstrated that large-scale motion and torque datasets generated by simplified-dynamics-based TO can provide effective priors for learning reusable locomotion skills and improving the efficiency of downstream RL~\cite{kang2026apt}. Motivated by this observation, ADP similarly constructs a TO-derived dataset, but uses the resulting dynamics distribution as a discriminator-defined prior for evaluating policy-generated temporal features, rather than as a trajectory reference, action decoder, footstep guide, or planning constraint.

\section{METHOD}
\label{section:method}
\subsection{Problem Formulation}

We train the humanoid locomotion policy with an asymmetric actor-critic formulation. The actor receives proprioceptive observations, including IMU signals, projected gravity, joint states, previous actions, and the velocity command. It outputs target position offset actions for the active joints. Base linear velocity is excluded from the actor input for real-world deployability, but is provided to the critic as privileged information during training.

The policy is trained to perform the locomotion task while regularizing its policy-induced features toward the TO-derived reference distribution. To this end, the total reward is defined as
\begin{equation}
r_t = w_G r_t^G + w_D r_t^D + w_R r_t^R,
\label{eq:total_reward}
\end{equation}
where $r_t^G$ is the task reward, $r_t^D$ is the proposed adversarial dynamics prior reward, and $r_t^R$ denotes regularization terms.

\subsection{Reference Trajectory Generation}
To construct the reference dataset for ADP training, we perform SRBD-based TO. We use SRBD as a low-dimensional TO model that represents the robot by its CoM dynamics and foot contact forces. The SRBD equations are defined as

\begin{subequations}
\label{eq:srbd}
\begin{equation}
    m \ddot{\mathbf{p}}^{\mathrm{com}}_t
    =
    m \mathbf{g}
    +
    \sum_{i \in \{L,R\}}
    \mathbf{f}^i_t,
\end{equation}
\begin{equation}
    \dot{\mathbf{L}}^{\mathrm{com}}_t
    =
    \sum_{i \in \{L,R\}}
    \left(
    \mathbf{p}^i_t - \mathbf{p}^{\mathrm{com}}_t
    \right)
    \times
    \mathbf{f}^i_t,
\end{equation}
\end{subequations}
where $m$ is the total robot mass, $\mathbf{p}^{\mathrm{com}}_t$ is the CoM position, $\mathbf{L}^{\mathrm{com}}_t$ is the centroidal angular momentum, $\mathbf{p}^i_t$ is the contact position of foot $i$, $\mathbf{f}^i_t$ is the corresponding contact force, and $i \in \{L, R\}$ denote the left and right foot contact points.

TO is used to generate reference locomotion trajectories from the SRBD model. Rather than serving as full-body motion references, the optimized trajectories provide SRBD-consistent feature samples over a range of commands. The TO problem is formulated as
\begin{equation}
\begin{aligned}
    \min_{\mathbf{x}_{0:T}, \mathbf{u}_{0:T}}
    \quad
    & \sum_{t=0}^{T}
    \left\|
    \mathbf{v}^{xy,\mathrm{com}}_t
    -
    \mathbf{v}^{xy,\mathrm{cmd}}
    \right\|^2_{\mathbf{Q}_v}
    +
    \left\|
    \mathbf{L}^{\mathrm{com}}_t
    \right\|^2_{\mathbf{Q}_L}
    +
    \sum_{i \in \{L,R\}}
    \left\|
    \mathbf{f}^{i}_t
    \right\|^2_{\mathbf{R}_f} \\
    \mathrm{s.t.}
    \quad
    & m \ddot{\mathbf{p}}^{\mathrm{com}}_t
    =
    m \mathbf{g}
    +
    \sum_{i \in \{L,R\}}
    \mathbf{f}^{i}_t, \\
    & \dot{\mathbf{L}}^{\mathrm{com}}_t
    =
    \sum_{i \in \{L,R\}}
    \left(
    \mathbf{p}^{i}_t - \mathbf{p}^{\mathrm{com}}_t
    \right)
    \times
    \mathbf{f}^{i}_t, \\
    & 0 \leq f^{i,z}_t \leq F_{\max}~\rho^{i}_t,\quad
    \sqrt{
    (f^{i,x}_t)^2 + (f^{i,y}_t)^2
    }
    \leq
    \mu f^{i,z}_t, \\
    & 0 \leq \rho^{i}_t \leq c^{i}_t,\quad c^{i}_t \in \{0,1\}.
\end{aligned}
\label{eq:srbd_to}
\end{equation}
Here, $\mathbf{x}_t$ includes the CoM state and centroidal momentum, while $\mathbf{u}_t$ includes the contact forces. The objective encourages the horizontal CoM velocity to follow the commanded velocity while penalizing excessive centroidal angular momentum and contact-force magnitudes. The constraints enforce SRBD consistency, contact-force activation, and friction-cone limits, producing trajectories that are physically grounded at the SRBD level.

For turning motions, the commanded yaw rate $\omega_z^{\mathrm{cmd}}$ is integrated into a time-varying heading frame in which the horizontal CoM velocity objective is evaluated; thus, no separate yaw-rate tracking term is required in \eqref{eq:srbd_to}. The TO trajectories are not used as full-body tracking targets; they only provide the reference set against which policy-generated windows are evaluated.

The dataset consists of multiple locomotion modes, including forward running, walking, turning, lateral stepping, and backward walking. Each motion is generated as an SRBD-consistent locomotion trajectory under a fixed linear and yaw velocity command, from which we extract dynamics quantities such as CoM motion, centroidal momentum, contact forces, and contact indicators. The contact schedule $\{c_t^L,c_t^R\}$ follows a per-motion prescribed gait schedule and is used as the binary contact indicator in each reference window. The continuous variable $\rho_t^i$ in \eqref{eq:srbd_to} only serves as a soft contact activation for scaling normal forces during optimization. Thus, $\mathcal{D}_{dyn}$ stores dynamics features rather than motion-tracking targets.
\subsection{Dynamics Feature Representation}
ADP compares TO-derived reference trajectories and policy rollouts in a shared feature space. To this end, we construct a dynamics feature at each timestep that includes CoM motion, centroidal momentum, contact forces, and contact states. This representation is separated from the actor observation and is used only as the discriminator input, not as the actor input.

The timestep-level dynamics feature is defined as
\begin{equation}
\begin{aligned}
\mathbf{z}_t =
[
&\mathbf{v}^{\mathrm{com}}_{t,h},\;
\mathbf{L}^{\mathrm{com}}_{t,h},\;
\bar{\dot{\mathbf{L}}}^{xy,\mathrm{com}}_{t,h},\;
\hat{\mathbf{f}}^{L}_{t,h},\;
\hat{\mathbf{f}}^{R}_{t,h},\;
c^{L}_t,\;
c^{R}_t
].
\end{aligned}
\label{eq:adp_feature_perstep}
\end{equation}
Here, the subscript $h$ denotes the heading frame. $\mathbf{v}^{\mathrm{com}}_{t,h}$ denotes the heading-frame CoM velocity, $\mathbf{L}^{\mathrm{com}}_{t,h}$ denotes the centroidal angular momentum, and $\dot{\mathbf{L}}^{xy,\mathrm{com}}_{t,h}$ denotes the planar angular momentum rate. In addition, $\hat{\mathbf{f}}^{L}_{t,h}$ and $\hat{\mathbf{f}}^{R}_{t,h}$ denote the left and right foot contact forces normalized by the robot weight, and $c^L_t,c^R_t$ denote binary contact indicators.

For policy rollouts, the centroidal angular momentum is computed by summing the link-wise spin momentum and the orbital momentum about the system CoM over all links of the articulated humanoid model:
\begin{equation}
\mathbf{L}^{\mathrm{com}}_t
=
\sum_i
\left(
\mathbf{I}_{i,w}\boldsymbol{\omega}_i
+
(\mathbf{p}_i-\mathbf{p}^{\mathrm{com}})
\times
m_i(\mathbf{v}_i-\mathbf{v}^{\mathrm{com}})
\right).
\end{equation}
This captures whole-body momentum redistribution in policy rollouts, while SRBD-based reference trajectories are converted into the same feature interface.

Because the same dynamics feature can correspond to different behaviors under different velocity commands, the discriminator input is conditioned on both the current command and the instantaneous tracking error. These quantities enter as conditioning variables that align the reference distribution with the current command, rather than as tracking-error reward terms. Given the command $\mathbf{v}^{\mathrm{cmd}}_t=[v^{\mathrm{cmd}}_{x,t}, v^{\mathrm{cmd}}_{y,t}, \omega^{\mathrm{cmd}}_{z,t}]$, the command-conditioned dynamics feature is defined as
\begin{equation}
\boldsymbol{\xi}_t
=
\left[
\mathbf{z}_t,\;
\mathbf{v}^{\mathrm{cmd}}_t,\;
\mathbf{e}_t
\right].
\label{eq:adp_feature_command}
\end{equation}
Here, $\mathbf{e}_t$ denotes the planar velocity and yaw-rate tracking error:
\begin{equation}
\mathbf{e}_t
=
\left[
v^{\mathrm{com}}_{x,t} - v^{\mathrm{cmd}}_{x,t},\;
v^{\mathrm{com}}_{y,t} - v^{\mathrm{cmd}}_{y,t},\;
\omega_{z,t} - \omega^{\mathrm{cmd}}_{z,t}
\right].
\end{equation}

Contact switching, force modulation, and momentum recovery cannot be sufficiently represented by a single timestep. Therefore, ADP constructs a short temporal window:
\begin{equation}
\boldsymbol{\Xi}_t
=
\left[
\bar{\boldsymbol{\xi}}_{t-K+1},
\bar{\boldsymbol{\xi}}_{t-K+2},
\ldots,
\bar{\boldsymbol{\xi}}_{t}
\right],
\end{equation}
where $\bar{\boldsymbol{\xi}}_t$ denotes the normalized dynamics feature and $K$ is the window length. The discriminator receives the flattened vector $\mathrm{vec}(\boldsymbol{\Xi}_t)$ as input.

Using this interface, TO-derived trajectories and policy rollouts are converted into reference and policy-generated windows, respectively. The reference dataset is defined as $\mathcal{D}_{\mathrm{dyn}} = \left\{ \boldsymbol{\Xi}^{\mathrm{ref}}_j \right\}_{j=1}^{N}.$  The corresponding rollout window at timestep $t$ is denoted by $\boldsymbol{\Xi}^{\pi}_t$. Although the reference windows are obtained from SRBD-based TO and the policy windows from articulated humanoid rollouts, both are represented in the same feature space. Therefore, ADP can compare the reference and policy-generated window distributions without using reference joint poses, motion phases, or end-effector trajectories. This comparison is intentionally defined in the selected feature space: the SRBD-derived reference provides SRBD-consistent features, but does not impose full-body joint-level kinematic targets or guarantee full-order dynamic feasibility of every articulated configuration.
\subsection{Training}

\begin{figure*}[t] 
\centering 
\includegraphics[width=\linewidth]{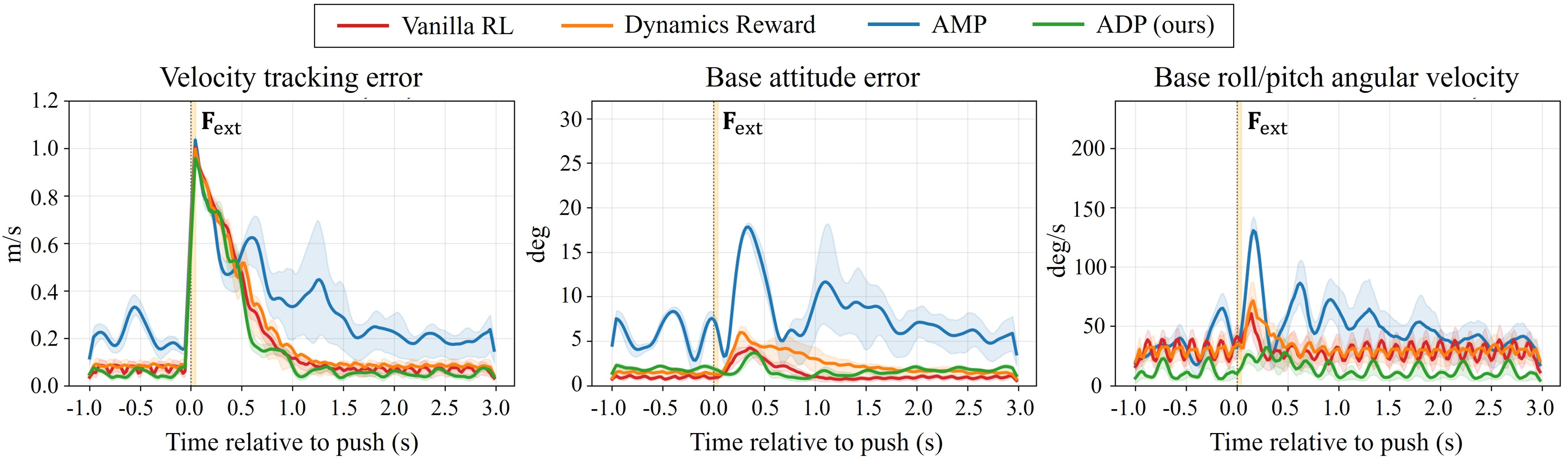} 
\caption{Time-series perturbation recovery under a lateral push. Red, orange, blue, and green curves denote Vanilla RL, Dynamics Reward, AMP, and ADP, respectively. Lines show the mean and translucent bands show the Interquartile Range (IQR) over 32 environments; the panels show velocity tracking error, base attitude error, and base roll/pitch angular velocity.
}
\label{fig:push_recovery_timeseries} 
\end{figure*}

We jointly perform PPO~\cite{schulman2017proximal}-based actor-critic learning and adversarial dynamics discriminator learning. During training, rollout windows $\boldsymbol{\Xi}^{\pi}_t$ are computed from policy rollouts and stored in a replay buffer $\mathcal{B}_{\pi}$. The discriminator is trained to distinguish rollout windows sampled from $\mathcal{B}_{\pi}$ from reference windows sampled from $\mathcal{D}_{\mathrm{dyn}}$.

The discriminator uses normalized windows as input, rather than single state transitions. It is optimized with the following least-squares adversarial objective:
\begin{equation}
\begin{aligned}
\mathcal{L}_{D}(\phi)
=
&\frac{1}{2}
\mathbb{E}_{\boldsymbol{\Xi}^{\mathrm{ref}}\sim\mathcal{D}_{\mathrm{dyn}}}
\left[
\left(
D_\phi(\boldsymbol{\Xi}^{\mathrm{ref}})-1
\right)^2
\right]
\\
&+
\frac{1}{2}
\mathbb{E}_{\boldsymbol{\Xi}^{\pi}\sim\mathcal{B}_{\pi}}
\left[
\left(
D_\phi(\boldsymbol{\Xi}^{\pi})+1
\right)^2
\right]
\\
&+
\lambda_{\mathrm{gp}}
\mathbb{E}_{\boldsymbol{\Xi}^{\mathrm{ref}}\sim\mathcal{D}_{\mathrm{dyn}}}
\left[
\left\|
\nabla_{\boldsymbol{\Xi}}
D_\phi(\boldsymbol{\Xi}^{\mathrm{ref}})
\right\|^2
\right].
\end{aligned}
\end{equation}
The first two terms classify reference and policy-generated windows as +1 and -1, respectively, while the gradient penalty stabilizes training around the reference distribution.

The policy receives the ADP reward from the discriminator output. For a policy-generated window $\boldsymbol{\Xi}^{\pi}_t$, the reward is defined as
\begin{equation}
r^D_t
=
c_D
\max
\left[
0,\;
1
-
\frac{1}{4}
\left(
D_\phi(\boldsymbol{\Xi}^{\pi}_t)-1
\right)^2
\right],
\label{eq:adp_reward}
\end{equation}
where $c_D$ is the ADP reward scale. This reward increases when the policy-generated window is classified as close to the reference distribution.

The actor and critic are updated using PPO with the total reward defined in~\eqref{eq:total_reward}. In contrast, the discriminator is updated using $\mathcal{L}_D$ with samples from $\mathcal{D}_{\mathrm{dyn}}$ and $\mathcal{B}_{\pi}$. The discriminator loss is not directly backpropagated through the simulator dynamics into the actor. Instead, the actor is affected by the dynamics prior only through the reward $r^D_t$. Thus, ADP affects the actor only through the reward $r_t^D$, without directly tracking reference joint poses, velocities, or end-effector trajectories. 
\section{EXPERIMENTS}
\label{section:experiments}
We quantitatively evaluate ADP in simulation and provide qualitative hardware demonstrations on the Unitree G1~\cite{unitree_g1}, a 29-DoF humanoid robot. The supplementary video provides side-by-side hardware comparison between ADP and AMP under manually applied external pushes. These hardware trials are intended as qualitative transfer validation rather than as a controlled quantitative evaluation. All methods share the same PPO implementation, architecture, observations, actions, command distribution, domain randomization, and perturbation curriculum. Auxiliary reward scales for non-vanilla methods are tuned using the same preliminary perturbation-recovery criterion and then fixed for all reported evaluations. The source code and experiment configurations will be released on the paper website. We compare ADP against Vanilla RL, which is trained only with task rewards without using a dataset; AMP~\cite{peng2021amp}, which learns a kinematic motion prior from reference data generated under the same locomotion set; and a Dynamics Reward baseline, which directly uses dynamics-feature matching rewards without an adversarial discriminator. The Dynamics Reward baseline uses the same feature set and reference data as ADP, but replaces the discriminator with a cluster-based matching reward
\begin{equation}
    r_t^{\mathrm{DR}}
    =
    \exp\!\left(
    -\tau \sum_{k} w_k(\mathbf{v}_t^{\mathrm{cmd}})\,
    \bigl\lVert
    (\boldsymbol{\Xi}^{\pi}_t - \boldsymbol{\mu}_k) \oslash \boldsymbol{\sigma}_k
    \bigr\rVert_2^2
    \right),
    \label{eq:dyn_reward}
\end{equation}
where $\{\boldsymbol{\mu}_k, \boldsymbol{\sigma}_k\}$ denote the per-feature mean and standard deviation of the $k$-th reference motion cluster, $\oslash$ denotes element-wise division, and $w_k(\mathbf{v}_t^{\mathrm{cmd}}) \propto \exp(-\lVert (\mathbf{v}_t^{\mathrm{cmd}} - \mathbf{c}_k) \oslash \boldsymbol{\sigma}_{\mathrm{cmd}}\rVert^2)$ weights the $k$-th reference motion cluster according to its command-space proximity to $\mathbf{v}_t^{\mathrm{cmd}}$. Here, $\mathbf{c}_k$ is the command center of the $k$-th cluster, and $\tau$ is a temperature. The Dynamics Reward thus measures a point-wise Gaussian-style distance from the policy window $\boldsymbol{\Xi}^{\pi}_t$ to the reference cluster centers, whereas ADP uses an adversarial discriminator to provide a distribution-level reward over reference windows. For a fair comparison, AMP and ADP use reference data generated from the same locomotion trajectories. This design controls the reference-source variable, so that the comparison isolates the effect of the prior representation rather than differences in demonstration quality. For AMP, following the TO-derived motion source used by Wu \textit{et al.}~\cite{wu2023learning}, these trajectories are converted into full-body kinematic motions through inverse kinematics, and the AMP discriminator receives joint-level kinematic features, including joint pose, joint velocity, and end-effector transforms. In contrast, ADP uses only command-conditioned windows extracted from the same trajectory sources, so that the structural difference between the two methods is the prior feature space rather than the reference source. The evaluation focuses on the following three questions:
\begin{itemize}
    \item \textbf{Q1:} Does ADP improve perturbation recovery performance after external disturbances?
    \item \textbf{Q2:} Does the proposed representation expose perturbation-induced transients more directly than a kinematic representation?
    \item \textbf{Q3:} Which dynamics features and temporal representations contribute to the performance improvement of ADP?
\end{itemize}
\subsection{Perturbation Recovery under External Disturbances}

We apply an instantaneous velocity impulse to the floating base in four directions (lateral $\pm y$, forward $+x$, backward $-x$) to assess generalization across disturbance directions. $J_{80}$ is determined by sweeping $\Delta v$ from 1.5 m/s to 4.5 m/s at 0.5 m/s intervals over the four push directions. Per-direction success rate, recovery time, and velocity tracking error are then measured under a unified push magnitude of $\Delta v = 3.0~\mathrm{m/s}$ (impulse $= 99~\mathrm{N{\cdot}s}$), and Table~\ref{tab:push_recovery_metrics} reports the corresponding direction-averaged metrics together with $J_{80}$.

The metrics in Table~\ref{tab:push_recovery_metrics} are defined as follows. \textbf{Success Rate} is the fraction of trials without fall termination during the post-push window, where a fall is declared if any of the simulator's termination conditions is triggered: contact at a designated termination body (the torso), base height below $0.30~\mathrm{m}$, or a lateral projected-gravity component (along the body $x$- or $y$-axis) exceeding $0.8$, which corresponds to an absolute pitch or roll of about $53^\circ$. \textbf{Recovery Time} measures how quickly the robot returns below fixed velocity and attitude error thresholds after the push, and \textbf{Velocity
Error} is the post-push RMS planar velocity tracking error. Fallen trials are assigned fixed penalties for recovery time and velocity error. These metrics are averaged over four push directions. \textbf{$J_{80}$} (the $80\%$-success impulse threshold) is the largest impulse $J = m\,\Delta v$ for which the direction-averaged success rate remains above $80\%$, identified by sweeping $\Delta v$ in $0.5~\mathrm{m/s}$
increments.

\begin{figure}[t] 
\centering 
\includegraphics[width=\linewidth]{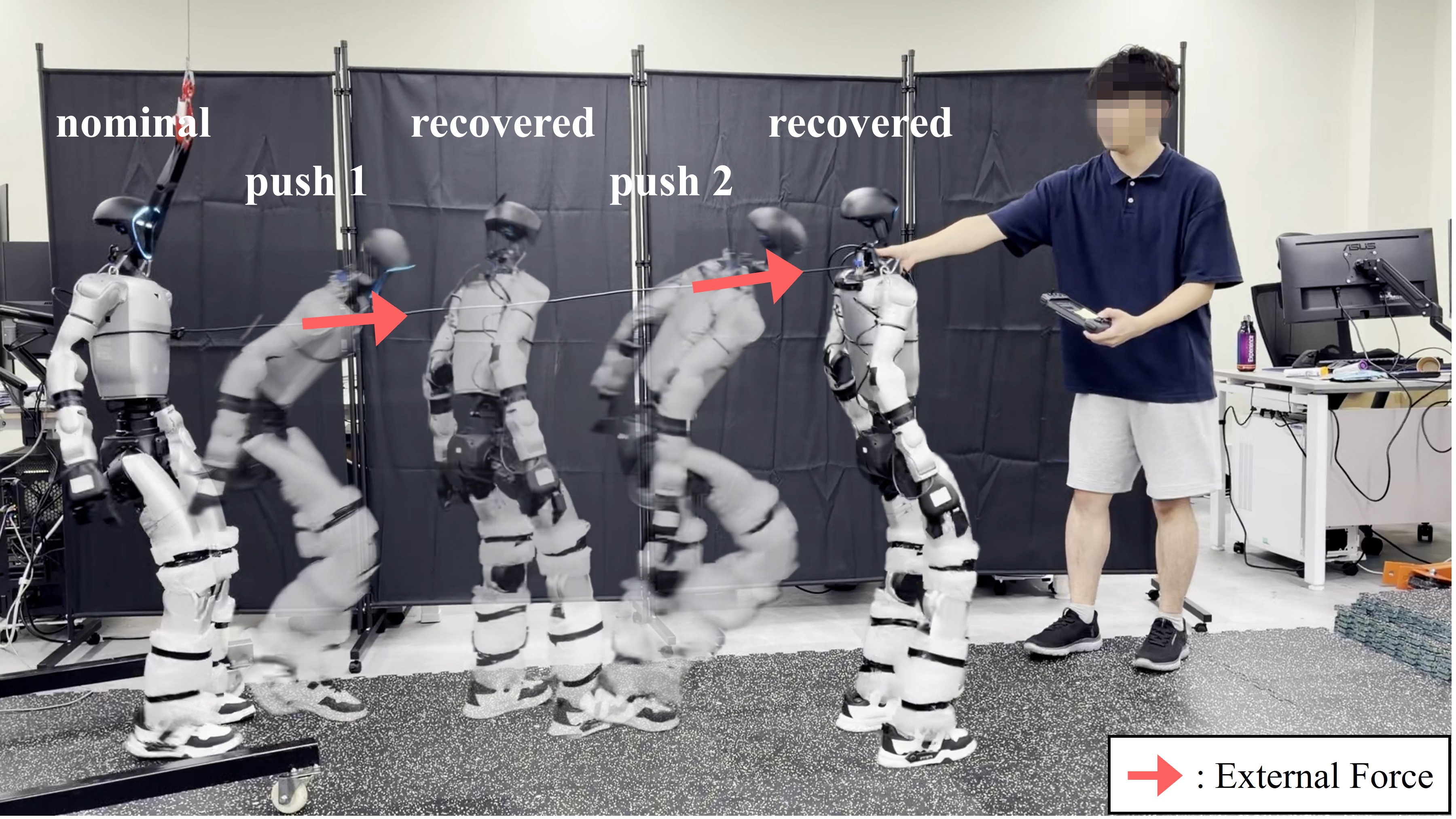} 
\caption{Qualitative hardware demonstration of ADP on the real robot under repeated external pushes. The overlaid snapshots show the robot transitioning from nominal walking to a perturbed state and recovering under subsequent perturbations.
}
\label{fig:hardware_push_recovery_snapshots} 
\end{figure}

Fig.~\ref{fig:push_recovery_timeseries} compares the recovery behaviors of the four policies under a representative lateral $\Delta v_y=3.0~\mathrm{m/s}$ push. In this representative lateral-push case, ADP rapidly attenuates the transient velocity and attitude errors within approximately $1$--$1.5~\mathrm{s}$, stabilizing the velocity error below $0.2~\mathrm{m/s}$ and the attitude error below $3^\circ$, while the formal direction-averaged recovery time is reported in Table~\ref{tab:push_recovery_metrics}. AMP recovers velocity tracking, but its base attitude oscillates near $10$--$15^\circ$ for almost $2~\mathrm{s}$, with angular velocity peaks exceeding $100^\circ/\mathrm{s}$. This suggests that kinematic-style matching can recover the visible posture in this case, but may leave under-damped residual rotational dynamics.

\begin{table}[t]
\centering
\caption{ Quantitative perturbation recovery performance under four-direction external pushes. }
\label{tab:push_recovery_metrics}
\scriptsize
\setlength{\tabcolsep}{2.2pt}
\renewcommand{\arraystretch}{1.15}
\resizebox{\columnwidth}{!}{
\begin{tabular}{lcccc}
\hline
\makecell[l]{\textbf{Method}} &
\makecell[c]{\textbf{Success}\\\textbf{Rate (\%)} $\uparrow$} &
\makecell[c]{$\boldsymbol{J_{80}}$\\\textbf{(N$\cdot$s)} $\uparrow$} &
\makecell[c]{\textbf{Recovery}\\\textbf{Time (s)} $\downarrow$} &
\makecell[c]{\textbf{Vel. Error}\\\textbf{(m/s)} $\downarrow$} \\
\hline
Vanilla RL &
5.5 &
49.5 &
10.52 &
2.89 \\
Dynamics Reward &
14.1 &
66.0 &
9.73 &
2.68 \\
AMP~\cite{peng2021amp} &
73.4 &
99.0 &
4.76 &
1.30 \\
ADP (ours) &
\textbf{91.4} &
\textbf{115.5} &
\textbf{2.48} &
\textbf{0.84} \\
\hline
\end{tabular}
}
\end{table}

Table~\ref{tab:push_recovery_metrics} quantitatively summarizes the perturbation recovery performance. ADP achieves the highest $J_{80}$ of $115.5$ N$\cdot$s and the highest direction-averaged success rate of $91.4\%$ across the four push directions. It also reduces direction-averaged recovery time by $47.9\%$ compared with AMP and by $74.5\%$ compared with Dynamics Reward, while achieving the lowest velocity-error, indicating that the recovery improvement holds across push directions rather than being specialized to a single direction.

The comparison with AMP highlights the benefit of regularizing selected dynamics features rather than only matching kinematic motion style, yielding a $48\%$ reduction in direction-averaged recovery time and a $35\%$ reduction in velocity error. The comparison with the Dynamics Reward baseline further isolates the effect of replacing point-wise feature matching with distribution-level adversarial regularization over the same feature set, suggesting that the adversarial formulation provides a more effective signal under this setup. Fig.~\ref{fig:hardware_push_recovery_snapshots} qualitatively demonstrates repeated external-push recovery on the real robot.

\subsection{Representation Sensitivity to External Perturbations}

We analyze representation sensitivity to external perturbations by comparing how strongly the same post-push transient appears in the proposed and kinematic representations.

For the same perturbed rollouts, we compute normalized deviations from each representation's own reference distribution. $d_{\rm dyn}(t)$ is computed from the dynamics feature $z_t$ in~\eqref{eq:adp_feature_perstep}, excluding the command and tracking-error variables in~\eqref{eq:adp_feature_command}. $d_{\rm kin}(t)$ is computed from the joint pose and joint velocity component of the AMP discriminator input. Each curve is normalized by its pre-push mean and computed from pooled non-fallen rollout episodes of all four methods under the same lateral-push condition.

\begin{figure}[t]
\centering
\includegraphics[width=\columnwidth]{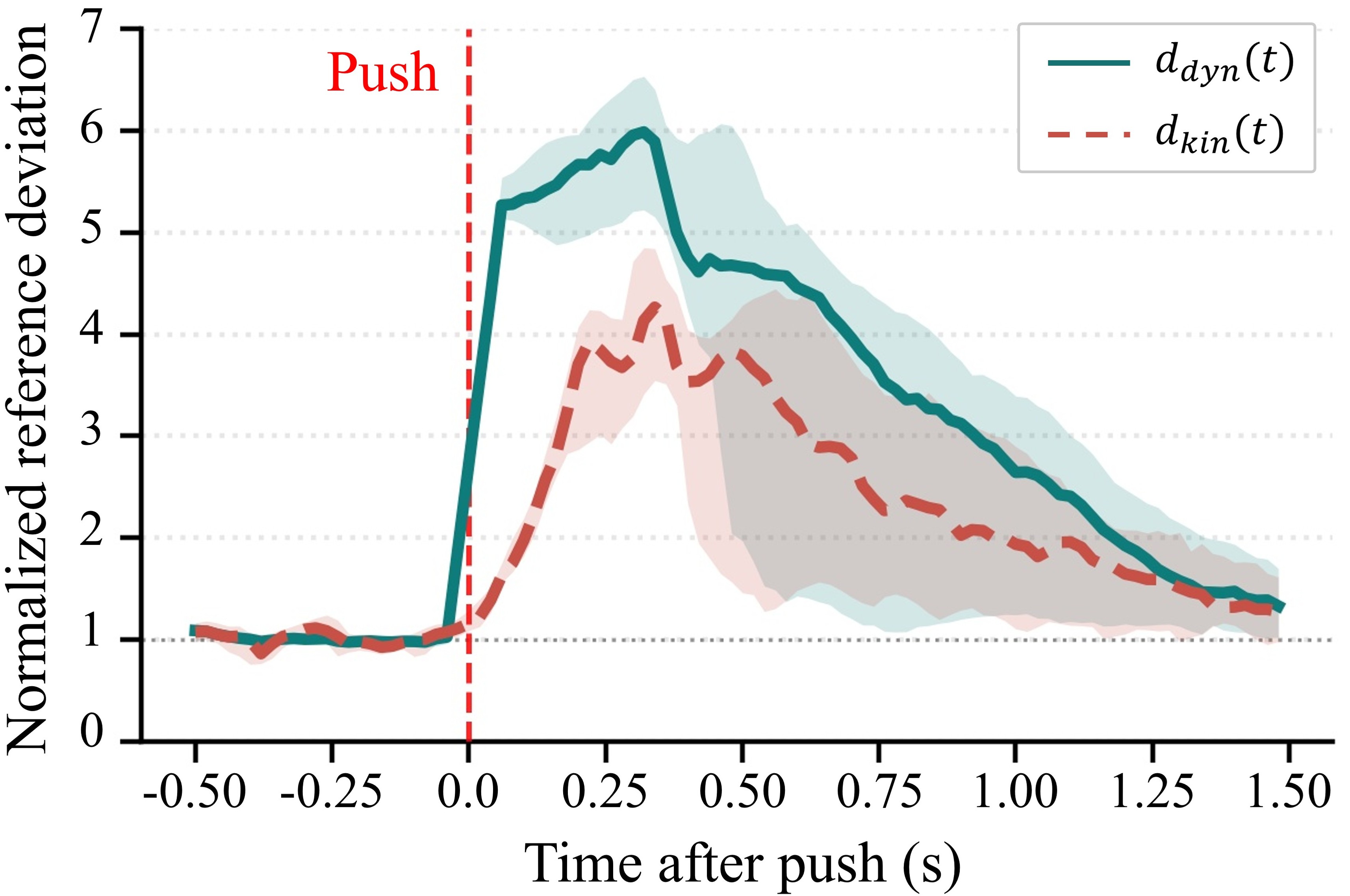}
\caption{Normalized reference deviation in dynamics-feature and kinematic-feature representations under a lateral $\Delta v_y=3.0$ m/s push. Lines and bands denote medians and IQRs over 106 pooled non-fallen rollout episodes.}
\label{fig:why_dynamics_prior}
\end{figure}

Fig.~\ref{fig:why_dynamics_prior} and Table~\ref{tab:priorspace_sensitivity} show that the proposed representation responds earlier and more strongly to the push: its deviation reaches $6.0\times$ the pre-push baseline within $20~\mathrm{ms}$, whereas the kinematic representation reaches $4.3\times$ with a $160~\mathrm{ms}$ rise time. The higher post-push Area Under the Curve (AUC) and pre/post separability further indicate that the push transient is more salient in the proposed representation. This is consistent with the fact that the impulse directly excites floating-base momentum and contact forces before its effect is expressed through joint-level kinematics.

This analysis is not a task-level performance metric and does not imply that raw deviation is monotonic with the discriminator reward in~\eqref{eq:adp_reward}. Fig.~\ref{fig:why_dynamics_prior} and Table~\ref{tab:priorspace_sensitivity} show that perturbation-induced transients are exposed earlier and more strongly in the proposed representation. These results indicate that impulse-induced recovery transients are more directly observable in the proposed feature space than in joint-level kinematic features.

\begin{table}[t]
\centering
\caption{Perturbation Sensitivity of the Dynamics-Feature and Kinematic-Feature Representations.}
\label{tab:priorspace_sensitivity}
\scriptsize
\setlength{\tabcolsep}{4.0pt}
\renewcommand{\arraystretch}{1.20}
\resizebox{\columnwidth}{!}{%
\begin{tabular}{lcccc}
\hline
\textbf{Representation} &
\makecell[c]{\textbf{Peak fold}\\\textbf{change} $\uparrow$} &
\makecell[c]{\textbf{50\%-rise}\\\textbf{time (ms)} $\downarrow$} &
\makecell[c]{\textbf{Post-push}\\\textbf{AUC} $\uparrow$} &
\makecell[c]{\textbf{Pre/post}\\\textbf{sep. AUC} $\uparrow$} \\
\hline
Kinematic &
$4.3\times$ & 160 & 2.2 & 0.98 \\
Dynamics &
$\mathbf{6.0\times}$ & $\mathbf{20}$ & $\mathbf{3.9}$ & $\mathbf{1.00}$ \\
\hline
\end{tabular}
}
\end{table}

\subsection{Ablation Study}
We next ablate the discriminator input while keeping the PPO setting, reward weights, domain randomization, and reference dataset fixed. All variants are evaluated under four-direction $\Delta v=3.0$ m/s pushes. In Table~\ref{tab:feature_component_ablation_metrics}, Dyn. Dist. denotes the post-push average $d_{\rm dyn}(t)$ from Sec. IV-B, with fallen episodes assigned $Z_{\rm pen}=5$.

Table~\ref{tab:feature_component_ablation_metrics} summarizes the component ablation results. Full ADP achieves the best performance across all metrics ($91.4\%$ success). Removing the centroidal angular momentum reduces the success rate to $45.3\%$ and raises the recovery time from $2.48~\mathrm{s}$ to $6.74~\mathrm{s}$, while removing the contact force feature drops the success rate more modestly to $82.0\%$. This indicates that centroidal momentum and foot contact forces both contribute to post-perturbation recovery, with momentum playing the larger role under direction-averaged disturbances. The largest degradation arises from removing the binary contact indicator, which drops the success rate to $30.5\%$ and increases the recovery time to $8.82~\mathrm{s}$, confirming that swing/stance transitions and contact timing are the most critical signals for alignment with the reference distribution.

\begin{table}[t]
\centering
\caption{Summary of dynamics feature component ablation results.}
\label{tab:feature_component_ablation_metrics}
\scriptsize
\setlength{\tabcolsep}{2.0pt}
\renewcommand{\arraystretch}{1.15}
\resizebox{\columnwidth}{!}{%
\begin{tabular}{lcccc}
\hline
\makecell[l]{\textbf{Method}} &
\makecell[c]{\textbf{Success}\\\textbf{Rate (\%)} $\uparrow$} &
\makecell[c]{\textbf{Vel. Error}\\\textbf{(m/s)} $\downarrow$} &
\makecell[c]{\textbf{Dyn.}\\\textbf{Dist.} $\downarrow$} &
\makecell[c]{\textbf{Recovery}\\\textbf{Time (s)} $\downarrow$} \\
\hline
ADP &
\textbf{91.4} &
\textbf{0.84} &
\textbf{1.13} &
\textbf{2.48} \\
ADP w/o Momentum &
45.3 &
1.92 &
3.04 &
6.74 \\
ADP w/o Contact Force &
82.0 &
1.06 &
1.51 &
3.32 \\
ADP w/o Contact Indicator &
30.5 &
2.36 &
3.82 &
8.82 \\
\hline
\end{tabular}%
}
\end{table}

\begin{figure}[t] 
\centering 
\includegraphics[width=\columnwidth]{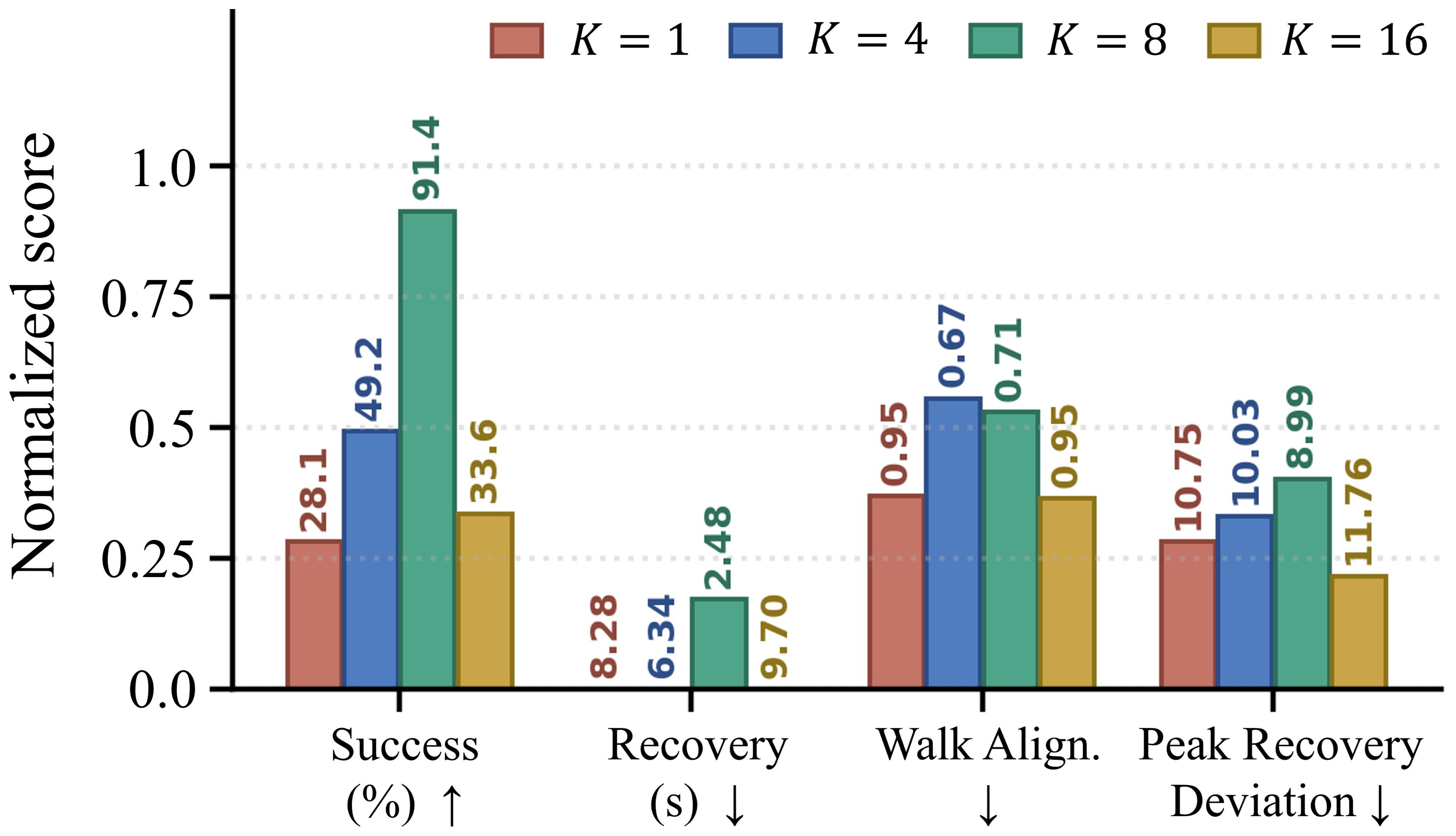} 
\caption{Effect of temporal dynamics-feature window length $K$ on normalized recovery metrics. Larger $K$ provides temporal context for contact switching, force modulation, and momentum recovery, but overly long windows can degrade training stability. Lower-is-better metrics are inverted after normalization, so higher scores indicate better performance. }
\label{fig:temporal_window_ablation} 
\end{figure}

Fig.~\ref{fig:temporal_window_ablation} shows the performance variation with respect to the temporal window length $K$, with all metrics direction-averaged. When $K=1$, the discriminator observes only single-timestep features and cannot sufficiently capture temporal patterns such as contact switching, force modulation, and momentum recovery, resulting in a direction-averaged success rate of only $28.1\%$. A moderate window length improves recovery by providing local temporal context, whereas an excessively large $K$ increases the discriminator input dimension and can overemphasize timing differences between SRBD-based reference trajectories and feedback-driven articulated rollouts. Although $K=4$ slightly improves walking-time alignment, it does not generalize consistently across push directions, resulting in a $49.2\%$ direction-averaged success rate, whereas $K=16$ further degrades to $33.6\%$ due to long-horizon timing mismatch and the increased discriminator input dimension. In contrast, $K=8$ achieves $91.4\%$ direction-averaged success with the lowest recovery time ($2.48\,\mathrm{s}$) and the smallest peak recovery deviation, providing the best trade-off between temporal context and training stability. We therefore adopt $K=8$ as the default temporal window for ADP.

\section{CONCLUSIONS}
\label{section:conclusion}
In this work, we proposed Adversarial Dynamics Priors (ADP) for physically grounded humanoid locomotion control. ADP provides an adversarial reward over TO-derived features instead of directly tracking reference motions, encouraging the policy to learn stable perturbation recovery behaviors. Experiments show that ADP reduces direction-averaged recovery time and velocity tracking error compared with AMP and direct dynamics-reward baselines. A complementary representation-sensitivity analysis further shows that the proposed feature space exposes perturbation-induced transients earlier and more strongly than a joint-level kinematic space. Since ADP does not explicitly enforce joint-level motion naturalness, future work will investigate combining dynamics and kinematic priors to improve both robustness and naturalness. We also note that ADP is compared against an AMP baseline using the same TO-derived reference source for fairness, rather than against a high-quality mocap-based AMP baseline; characterizing the gap between TO-derived and high-quality mocap-based kinematic priors is left to future work.








\bibliographystyle{ieeetr}

\end{document}